\documentclass[11pt]{article}
\usepackage{url}

\begin{document}
\title{Algorithmic Programming Language Identification}
\author{David Klein, Kyle Murray and Simon Weber\\
University of Rochester}
\date{November 1, 2011}
\maketitle

\begin{abstract}
Motivated by the amount of code that goes unidentified on the web, we introduce a practical method for algorithmically identifying the programming language of source code. Our work is based on supervised learning and intelligent statistical features. We also explored, but abandoned, a grammatical approach. In testing, our implementation greatly outperforms that of an existing tool that relies on a Bayesian classifier. Code is written in Python and available under an MIT license.
\end{abstract}

\newpage

\section{Introduction and Motivation}
The purpose of algorithmic programming language detection is to determine the programming language with which a particular program or program fragment was written. The goal of this research is to devise techniques to recognize and accurately identify code from as many programming languages as possible, without starting with any information other than the code itself.

One large domain in which many intensely useful programs and code fragments go unidentified is the web. Such code appears in blogs, forums, mailing lists, documentation, and many other contexts where the supporting web software does not provide a facility for identifying the language of a particular program. With identifying information, various possibilities are unlocked. For instance, code found on the web now becomes more searchable, more so than the limited form of searchability that tools such as Google's code search provide for explicitly marked full programs. In addition, code fragments could become more readable when they are automatically syntax highlighted based on the proper language grammar. Other potential uses range from data recovery on damaged file systems to enabling automatic code grading systems to be language agnostic.

\section{Existing Tools}
Simple tools for identifying source code already exist. Various syntax highlighting tools such as Google Code Prettify will automatically highlight syntax given some code \cite{GCP}. However, these tools do not actually identify languages; instead, they use heuristics that will make the highlighting work well. In the case of Google Code Prettify, broad grammars (such as C-like, Bash-like, and Xml-like) are preprogrammed. These grammars are then used to scan code, and the best matching grammar is used in highlighting. Clearly, languages that share a grammar cannot be distinguished between (and they do not need to be for the highlighting to work).

More relevant is SourceClassifier, which will attempt to identify a programming language given some code \cite{SC}. However, it relies on a simple Bayesian classifier. Its strength is therefore limited to the quality of training data, and it can easily be thrown off by strings and comments. These existing tools show a practical need for our work, and we are confident that they can be improved upon.

\section{Approach}
Our approach involves obtaining a large database of source code, written in a multitude of languages.  We then train our program on the database by giving it pieces of source code in identified languages.  We can use the resulting knowledge to evaluate source code from unknown languages and determine which language it is likely to be written in.

\subsection{Training Data}
All training data was collected from the code hosting website Github by a custom web crawler. Github's code tagging was used to initially identify samples. However, this was unlikely to be accurate in all cases. To address this, we verified the samples via file extension; only files that matched the common file extension for a language were used to train that language. This minimized noise in our training data.

Our training data is 324 megabytes and contains over 41 thousand source code files in the following languages:

\begin{itemize}
\item Actionscript
\item Ada
\item Brainfuck
\item C
\item C\#
\item C++
\item Common Lisp
\item CSS
\item Erlang
\item Haskell
\item HTML
\item Java
\item Javascript
\item Lua
\item Matlab
\item Objective C
\item Perl
\item PHP
\item Python
\item Ruby
\item Scala
\item Scheme
\item Smalltalk
\item Latex
\item XML
\end{itemize}

\subsubsection{Comment and String Detection}
Since comments and strings in any language can have arbitrary content, it is important to exclude their content from consideration during training and identification. Our algorithm for identifying comments and strings depends on a simple heuristic: comments and strings are some of the only parts of code that will contain natural language, and so will often have alphabetic-only words separated by spaces. We will define a line of code to have the \emph{words property} if it contains a sequence of alphabetic-only words separated by spaces and ending in a space or a newline. This is used in combination with other heuristics throughout the algorithm.

The first step in our algorithm identifies lines that are likely to contain a comment or string. This is done by simply finding lines that match the words property; these will be referred to as \emph{candidate lines}. For all candidate lines, in addition to the line number, the location of the string that matched the words property is also stored. This will be called the \emph{capture}. If a candidate line has more than one capture, the capture with the largest number of words is chosen. This is intended to guide the algorithm towards the most likely choices.

The algorithm performs the same general operations for each search type. A search is performed from current candidate lines for likely matches. During the search, the number of times that tokens or token combinations occur is recorded. After candidate lines are exhausted, lines that produced a match are removed from consideration on later searches. Finally, the most commonly occurring token or token combination is returned for each. The details of each search are as follows:

\begin{itemize}
\item Strings: A match has identical string tokens surrounding it on the same line. String tokens are read from a hardcoded database. This makes two assumptions: that string tokens are the same on either side and that string tokens will be in our database.

\item Block Comments: While lines maintain the words property, a search for opening tokens proceeds left and up from a capture. One additional line is searched when the words property no longer holds, since it is common to find an opening token on a line by itself. Each line is first split into whitespace separated tokens. Each token is then scored for its likelihood of being a starting block token. This is backed by a hardcoded database of common block comment tokens along with a heuristic to allow flexibility for tokens not hardcoded. The heuristic is based on a number of simple features, such as the occurrence of various characters and similarity to known tokens. If a token is thought to be likely, it is stored along with its position.

Then, for each opening token that was considered likely, a search for a matching closing token proceeds down and right. Closing tokens are scored using a similar heuristic, which is based on the symmetric nature of most block tokens.

After all likely block comments have been identified, a few extra operations are performed to attempt to reduce bad matches. Only the outermost comment block is considered in the case of nested comments. Nested comments were a rarity in testing, and tended to be bad matches. In addition, comments subsumed by another comment are removed if their tokens differ. This will handle cases where programmers delimit the interior of a block comment with a different token than the actual block tokens; this is common with ``*'' in C. Removing the ``*'' occurrences will boost the consideration of the actual block tokens.

\item Line Comments: Each remaining candidate line has the largest non-alphanumeric string at its left stripped of whitespace and considered as a line comment token. After all candidate lines have been considered, comment tokens are combined into the smallest possible token. For example,  tokens ``;;'', ``;('', and ``;'' will all combine to ``;''.
\end{itemize}

Naturally, a heuristics-based algorithm does not guarantee success. However, even if incorrect tokens are returned, the words property tends to guide the algorithm toward lines we would want to ignore, regardless of type of comment (or even whether or not the line was a comment). Formal tests were not run on the algorithm. However, informal testing showed that it was accurate across a range of random source files from our database. 

\subsection{Statistical Analysis}
The statistical aspects of our algorithm are based on several specific features.  We evaluate pieces of source code on those features and compare the results to those of each language in our database.  The features that we are using are as follows:

\begin{itemize}
\item Brackets: We evaluate code based on the relative prevalence of the different kinds of brackets: parentheses, curly braces, square brackets, and angle brackets.  This feature allows us to distinguish between languages like C and Java and languages like Lisp.
\item FirstWord: We look at the first word of each line of code.  A word is defined as a string of characters not separated by whitespace, including punctuation, letters and numbers.  This feature is intended to find common prefixes, like 'int' and 'public'.
\item Keywords: In this feature, we examine each individual word in the source code.  These words contain only letters and are separated by whitespace, numbers or punctuation.
\item LastCharacter: Here, we consider the final character of each line of code.  We can use this to identify languages where lines commonly end in specific characters, for example Prolog or Java.
\item Operators: Here, we strip out letters and numbers and attempt to identify operators by examining strings of punctuation delimited by spaces.
\item Punctuation: We compare the frequency of punctuation characters with the frequency of letters.  This is useful for identifying languages like the esoteric languages Chef and Brainfuck.
\item Comments and Strings: As described previously, we can often determine the tokens used to specify strings and comments in source code of an unknown language and see if the tokens are the same in known languages.
\end{itemize}

For each language and each feature, we calculate a score $s_l$.  The higher the score, the more likely it is that the language is the correct match for the source code.

We process each feature to obtain a numerical result representing the closeness that the source code has to each language in terms of that feature. The systems used for the features are as follows:

\begin{itemize}
\item FirstWord, Keywords, LastCharacter and Operators:
\begin{center}
$s_l = \frac{1}{\displaystyle\sum\limits_{i=1}^n (\frac{p_{i,l} - x_i}{p_{i,l}})^2}$\\
\end{center}
$p_{i,l}$ is the percentage of lines (or words or operators) that the $i$th most common character (or word or operator) in language $l$ appears in $l$.
$x_i$ is the percentage of lines (or words or operators) that the $i$th most common character (or word or operator) in language $l$ appears in in the source code in an unknown language.  We compensate for situations where some languages don't have enough different characters (or words or operators) for $i$ to reach $n$.
$n$ is a constant that varies according to the feature.

\item Brackets:
\begin{center}
$s_l = \frac{1}{\displaystyle\sum (p_{i,l} - x_i)^2}$\\
\end{center}
$p_{i,l}$ is the percentage of brackets in language $l$ that are of type $i$.
$x_i$ is the percentage of brackets in the unknown source code that are of type $i$.

\item Comments and Strings: $s_l$ is the simply the number of line comment tokens, block comment tokens, and string tokens in language $l$ that match exactly to a token in the unknown source code.

\item Punctuation: 
\begin{center}
$s_l = \frac{1}{\sqrt{\displaystyle\sum \|a - b_l\| + \|x - y_l\|}}$\\
\end{center}
$a$ is the ratio of the number of punctuation marks in the unknown source code to the number of punctuation marks added to the number of letters in the unknown source code.
$b_l$ is the ratio of the number of punctuation marks in language $l$ to the number of punctuation marks added to the number of letters in language $l$.
$x$ is the ratio of the number of letters in the unknown source code to the number of punctuation marks added to the number of letters in the unknown source code.
$y_l$ is the ratio of the number of letters in language $l$ to the number of punctuation marks added to the number of letters in language $l$.
\end{itemize}

We normalize all the $s_l$ values so that $\displaystyle\sum s_l$ for each feature equals 1.  Then we add the scores for each language together and compare them to each other to find the language that is most likely to be correct.

\subsection{Grammatical Approach}
To attempt to add another dimension to our identification, we also explored a grammar-based approach. Having ruled out learning grammars from scratch as too difficult and unlikely to produce accurate results, we decided to start with basic grammars for different features of broad kinds of languages (a C-style for loop would be an example, as would a ``new'' definition in Java). From here, when learning from a new source file, we would attempt to match each piece of grammar against the file. Those that matched well would be combined into a general grammar for that language, which could then help identify it in the future.

OMeta, an object-oriented language for pattern matching, was explored to facilitate this in code \cite{OMeta}. This project is based on parsing expression grammars, which are similar to context free grammars, but utilize prioritized choice to eliminate ambiguity when parsing. This allows a linear time parser for all PEGs. Two main problems had to be addressed for our high-level approach to work.

\subsubsection{Scoring a Grammar to Code}
First, for a piece of grammar to be useful to us, we had to have a way to score how well it fit some piece of code. Measuring the amount of code covered by parsing would be a simple way to do this. However, using actual lengths of tokens as a measurement would be inaccurate; a better method would be to count the number of tokens matched. This would attempt to even the importance of different grammatical pieces, since including token lengths could unfairly weight items with lengthy natural language tokens, such as variable declarations or comments.

\subsubsection{Combining Grammars}
The next step was to take two grammar pieces and combine them into a single grammar. OMeta's ability to switch between grammars easily would have allowed us to compose a grammar just from an ordering of the pieces. However, without knowledge of the priority of the ordering, a PEG would be difficult to construct.

Despite our efforts in this area, we were unable to combine this approach into our main framework. Both finding grammar pieces and combining grammars proved to be barriers to implementation. We leave this approach to future work.

\section{Results}
We tested our algorithm on 25 randomly selected source code files from our database. Note that we do not examine the file extension as a part of our identification, since we are intending this as a simulation of the uses described earlier. Our program ranks languages from most likely to least likely.  Here are the results of this test:

\begin{itemize}
\item Twelve out of the 25 files (48\%) were identified correctly.
\item Four times (12\%), the algorithm chose the correct language as the second most likely language.
\item Four times (12\%), the algorithm chose the correct language as the third most likely language.
\item Once (4\%), the algorithm chose the correct language as the fourth most likely language.
\item Four times (12\%), the algorithm did not identify the correct language as one of its top five languages.
\end{itemize}

Note that we were selecting the correct language out of a total of 25 known languages.  We consider this a promising start, especially considering that many of the languages are very similar (Scheme and Lisp, for example). 

For comparison, we ran a head-to-head test against a known tool, SourceClassifier. For fairness, we trained SourceClassifier on the same database we used to train our implementation. Here are the results on a different set of 25 files:

\begin{itemize}
\item Thirteen out of the 25 files (52\%) were identified correctly.
\item Two times (8\%), the algorithm chose the correct language as the second most likely language.
\item Six times (24\%), the algorithm chose the correct language as the third most likely language.
\item Two times (8\%), the algorithm chose the correct language as the fourth most likely language.
\item Two times (8\%), the algorithm did not identify the correct language as one of its top five languages.
\end{itemize}

SourceClassifier performed much worse, identifying only four of the 25 files (12\%) correctly. This clearly shows the utility of our targeted methods over a simple bag of words approach.

\section{Code}
Python code is available on GitHub under an MIT license \cite{Code}. It was written quickly for a tight deadline, and has not been refactored since; this is intended to allow others to reproduce our results. The database used in testing is included.

\section{Future Work}
Various aspects of our algorithm and research direction are amenable to further study and innovation. Certainly, the grammatical approach can be explored more fully in order to compare its efficacy to that of the statistical approach, and to see if the two can be combined to form an even more accurate system. Although we explored a wide variety of statistical features in our analysis, even more remain to be studied. Because many interesting applications depend on the language identification features that we have explored, an exciting direction would be to embed our system in an application that makes use of our system in a practical and user friendly way.

\bibliographystyle{plain}
\bibliography{citations}

\end{document}